\pdfoutput=1

\documentclass[11pt]{article}

\usepackage{ACL2023}

\usepackage{times}
\usepackage{latexsym}

\usepackage[T1]{fontenc}

\usepackage[utf8]{inputenc}

\usepackage{microtype}

\usepackage{inconsolata}

\usepackage{xcolor}
\usepackage{booktabs, array}
\usepackage{tabularx}
\usepackage{hyperref}
\usepackage{graphicx}
\usepackage{adjustbox}
\usepackage{amsmath}
\usepackage{xspace}
\usepackage{multirow}
\usepackage{amssymb}
\usepackage{pifont}

\renewcommand{\eqref}[1]{Eq.~\ref{#1}}

\newcommand{\Method}{Distilling step-by-step\xspace}

\title{
Distilling Step-by-Step! Outperforming Larger Language Models\\
with Less Training Data and Smaller Model Sizes
}

\author{
Cheng-Yu Hsieh$^{1}$\thanks{\ \ Work done while the author was a student researcher at Google Cloud AI Research.}, \
Chun-Liang Li$^{2}$, \
Chih-Kuan Yeh$^{3}$, \
Hootan Nakhost$^{2}$, \\
\bf
Yasuhisa Fujii$^{3}$, \
Alexander Ratner$^{1}$, \
Ranjay Krishna$^{1}$, \
Chen-Yu Lee$^{2}$, \
Tomas Pfister$^{2}$ \\
$^1$University of Washington,
$^2$Google Cloud AI Research,
$^3$Google Research\\
\texttt{cydhsieh@cs.washington.edu}
}

\begin{document}
\maketitle
\begin{abstract}
Deploying large language models (LLMs) is challenging because they are memory inefficient and compute-intensive for practical applications. In reaction, researchers train smaller task-specific models by either finetuning with human labels or distilling using LLM-generated labels. However, finetuning and distillation require large amounts of training data to achieve comparable performance to LLMs. We introduce \emph{Distilling step-by-step}, a new mechanism that (a) trains smaller models that outperform LLMs, and (b) achieves so by leveraging less training data needed by finetuning or distillation. Our method extracts LLM rationales as additional supervision for training small models within a multi-task framework. We present three findings across $4$ NLP benchmarks: First, compared to both finetuning and distillation, our mechanism achieves better performance with much fewer labeled/unlabeled training examples. Second, compared to few-shot prompted LLMs, we achieve better performance using substantially smaller model sizes. Third, we reduce both the model size and the amount of data required to outperform LLMs; our finetuned 770M T5 model outperforms the few-shot prompted 540B PaLM model using only $80\%$ of available data on a benchmark, whereas standard finetuning the same T5 model struggles to match even by using $100\%$ of the dataset.\footnote{Source code is available at: \url{https://github.com/google-research/distilling-step-by-step}.}
\end{abstract}

\begin{figure}[!t]
    \centering
    \includegraphics[width=0.95\linewidth]{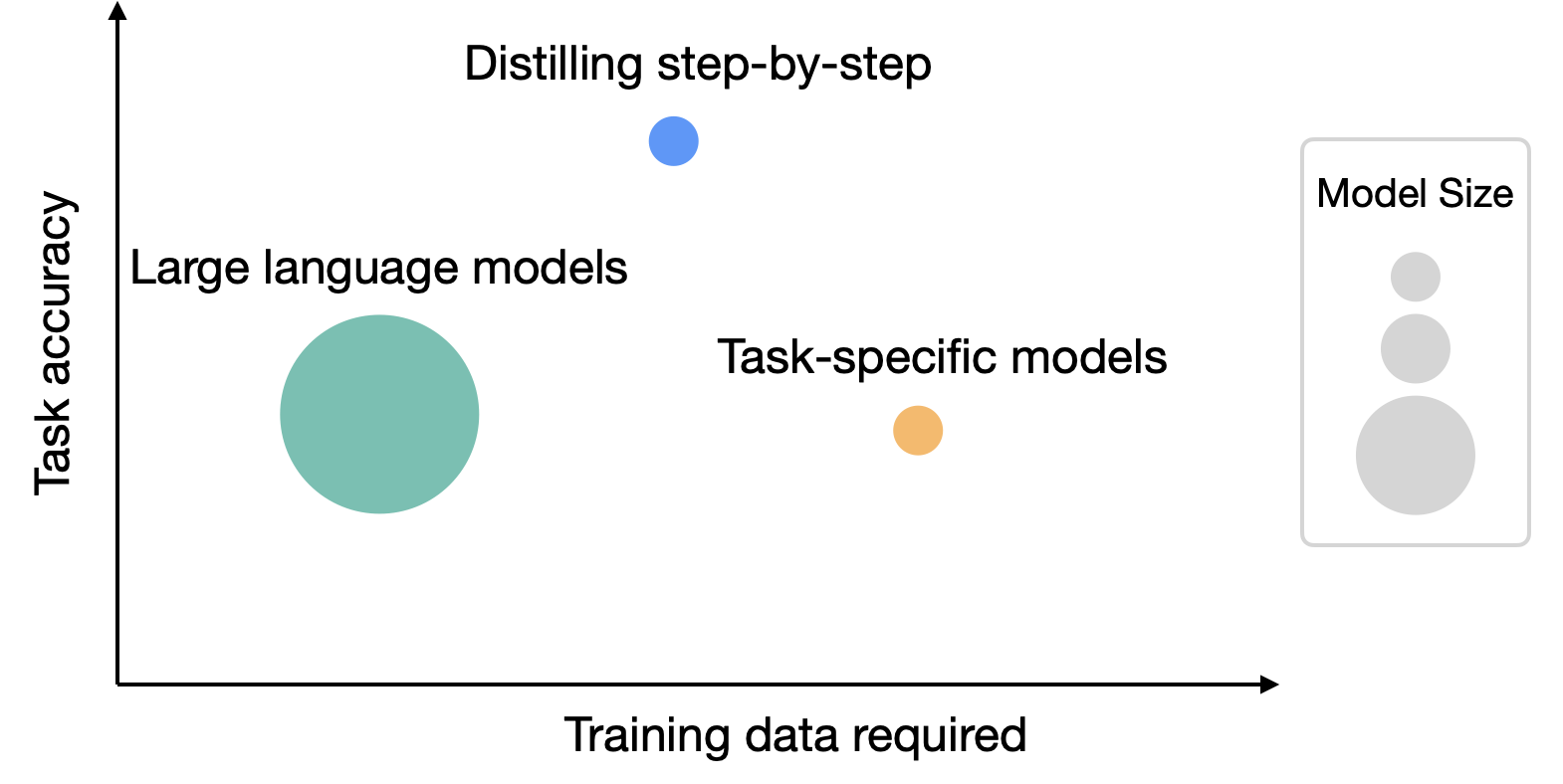}
    \vspace{-3mm}
    \caption{While large language models (LLMs) offer strong zero/few-shot performance, they are challenging to serve in practice. 
    Traditional ways of training small task-specific models, on the other hand, requires large amount of training data.
    We propose \Method, a new paradigm that extracts rationales from LLMs as informative task knowledge into training small models, which reduces both the deployed model size as well as the data required for training.}
    \label{fig:paradigm}
\end{figure}

\section{Introduction}
Despite the impressive few-shot ability offered by large language models (LLMs)~\citep{brown2020language,chowdhery2022palm,thoppilan2022lamda,hoffmann2022training,smith2022using,zhang2022opt}, these models are challenging to deploy in real world applications due to their sheer size. 
Serving a single $175$ billion LLM requires at least $350$GB GPU memory using specialized infrastructure~\citep{zheng2022alpa}. 
To make matters worse, today's state-of-the-art LLMs are composed of over $500$B parameters~\cite{chowdhery2022palm}, requiring significantly more memory and compute. 
Such computational requirements are far beyond affordable for most product teams, especially for applications that require low latency performance. 

To circumvent these deployment challenges of large models, practitioners often choose to deploy smaller specialized models instead. 
These smaller models are trained using one of two common paradigms: \textit{finetuning} or  \textit{distillation}. 
Finetuning updates a pretrained smaller model (e.g.~ BERT~\citep{devlin2018bert} or T5~\citep{raffel2020exploring}) using downstream human annotated data~\citep{howard-ruder-2018-universal}. 
Distillation trains the same smaller models with labels generated by a larger LLM~\citep{tang2019distilling,wang2021want,smith2022language,arora2022ask}.
Unfortunately, these paradigms reduce model size at a cost: to achieve comparable performance to LLMs, finetuning requires expensive human labels, and distillation requires large amounts of unlabeled data which can be hard to obtain~\citep{tang2019distilling,liang2020mixkd}.

In this work, we introduce \textbf{\Method}, a new simple mechanism for training smaller models with less training data. 
Our mechanism reduces the amount of training data required for both finetuning and distillation of LLMs into smaller model sizes.
Core to our mechanism is changing our perspective from viewing LLMs as a source of noisy labels to viewing them as agents that can reason: LLMs can produce natural language rationales justifying their predicted labels~\citep{wei2022chain,kojima2022large}. 
For example, when asked ``\textit{Jesse's room is $11$ feet long and $15$ feet wide. If she already has $16$ square feet of carpet. How much more carpet does she need to cover the whole floor?}'', an LLM can be prompted by chain-of-thought (CoT) technique~\citep{wei2022chain} to provide intermediate rationales ``\textit{Area $=$ length $\times$ width. Jesse’s room has $11\times15$ square feet.}'' that better connects the input to the final answer ``\textit{$( 11 \times 15 ) - 16$}''.
These \textit{rationales} can contain relevant task knowledge, such as ``\textit{Area $=$ length $\times$ width}'', that may originally require many data for small task-specific models to learn.
We thus utilize these extracted rationales as additional, richer information to train small models through a multi-task training setup, with both label prediction and rationale prediction tasks~\citep{raffel2020exploring,narang2020wt5}.

\Method allows us to learn task-specific smaller models that outperform LLMs using over $500\times$ less model parameters, and it does so with far fewer training examples compared to traditional finetuning or distillation (Figure~\ref{fig:paradigm}). 
Our results show three promising empirical conclusions across $4$ NLP benchmarks.
First, compared to both finetuning and distillation, our resulting models achieve better performance with over $50\%$ less training examples on average across datasets (and up to over $85\%$ reduction).
Second, our models outperform LLMs with much smaller model sizes (up to $2000\times$ smaller), drastically reducing the computation cost required for model deployment. 
Third, we simultaneously reduce the model size as well as the amount of data required to outperform LLMs.
We surpass the performance of $540$B parameter LLMs using a $770$M T5 model; this smaller model only uses $80\%$ of a labeled dataset that would otherwise be required if using an existing finetuning method. 
When only unlabeled data is present, our small models still perform on par or better than LLMs. We outperform $540$B PaLM's performance with only a $11$B T5 model.
We further show that when a smaller model performs worse than an LLM, \Method can more efficiently leverage additional unlabeled data to match the LLM performance compared to the standard distillation approach.
\begin{figure*}
    \centering
    \includegraphics[width=\linewidth]{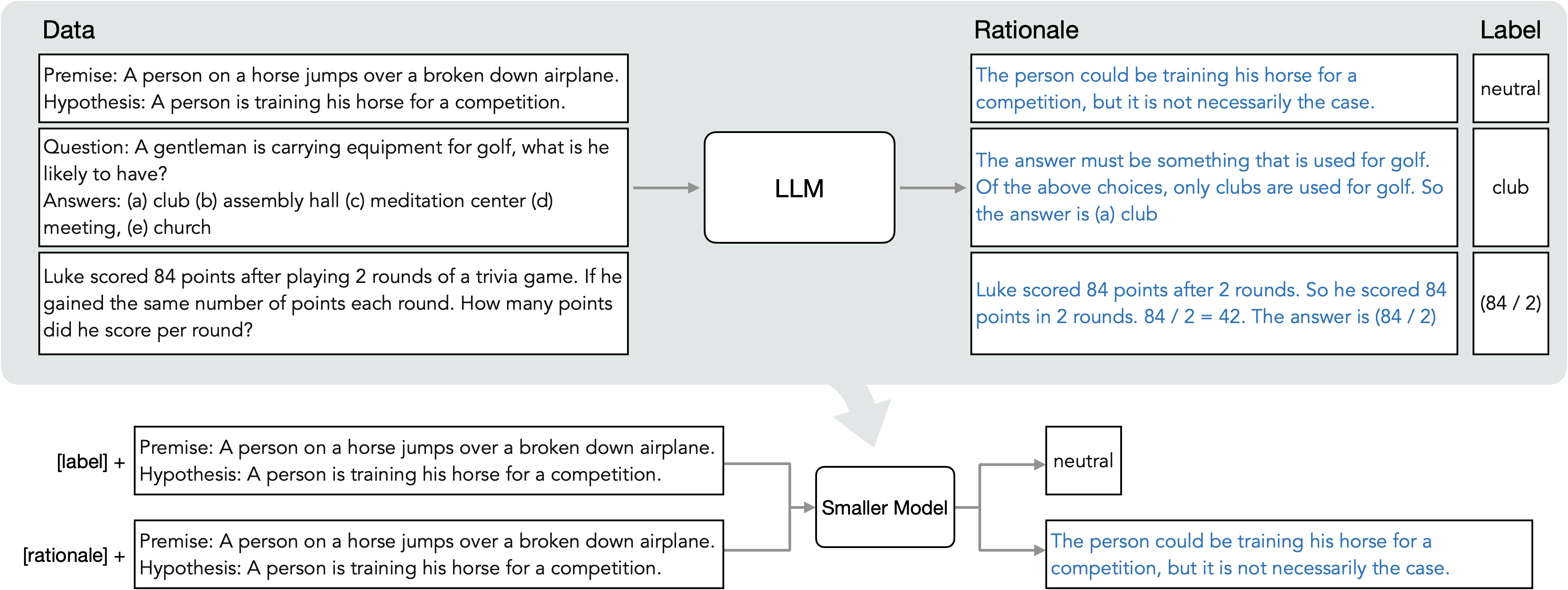}
    \caption{Overview on Distilling step-by-step. We first utilize CoT prompting to extract rationales from an LLM (Section~\ref{sec:rationale-extraction}). We then use the generated rationales to train small task-specific models within a multi-task learning framework where we prepend task prefixes to the input examples and train the model to output differently based on the given task prefix (Section~\ref{sec:multitask-training}).}
    \label{fig:framework}
\end{figure*}

\section{Related work}
Our work distills task-specific knowledge of LLMs into smaller specialist models by leveraging the emergent reasoning capabilities of today's LLMs. We draw on knowledge distillation research and methods that learn from both human-generated rationales and LLM-generated rationales.

\paragraph{Knowledge distillation from large models.}
Knowledge distillation has been successfully used to transfer knowledge from larger, more competent teacher models into smaller student models affordable for practical applications~\citep{bucilua2006model,hinton2015distilling,beyer2022knowledge,west2021symbolic,fu2023specializing}. It supports learning from limited labeled data, since the larger teacher model is often used to generate a training dataset with noisy pseudo labels~\citep{chen2020big,iliopoulos2022weighted,wang2021want,smith2022language,arora2022ask,agrawal2022qameleon}.
The one limitation that knowledge distillation often faces is its reliance on large amounts of unlabelled data required to create a useful noisy training dataset. Although prior work has explored using data augmentation techniques to reduce this hunger for data~\citep{tang2019distilling,liang2020mixkd,srinivas2018knowledge,milli2019model}, we propose an alternative approach: we reduce the need for large unlabeled data by distilling not just labels but also the teacher's rationales.

\paragraph{Learning with human rationales.}
While utilizing LLM-generated rationales is a new exciting area of investigation, using human-generated rationales has a rich history~\citep{hase2021can}. For instance, human rationales can be used to regularize model behavior~\citep{ross2017right}; it can be used as additional inputs to guide a model's predictions~\citep{rajani2019explain}; it can be used to improve overall model performance~\citep{zaidan-etal-2007-using,zhang-etal-2016-rationale,camburu2018snli,hancock2019learning,pruthi2022evaluating}; and human rationales can be used as gold standard labels to make models more interpretable by generating similar rationales~\citep{wiegreffe-etal-2021-measuring,narang2020wt5,eisenstein2022honest}.
Unfortunately, human rationales are expensive.

\paragraph{Learning with LLM generated rationales.}
Today's LLMs are capable of explaining their predictions by generating high-quality reasoning steps~\citep{wei2022chain,kojima2022large}. These reasoning steps have been used to augment input prompts to LLMs, improving their few-shot or zero-shot performance~\citep{wei2022chain,kojima2022large,wang2022self}; reasoning steps have also been used as additional finetuning data ``self-improve'' LLMs~\citep{zelikman2022star,huang2022large}.
Unfortunately, regardless of how LLMs are improved, their large size limits their utility in most test-time applications.

By contrast, we leverage generated rationales as informative supervision to train smaller task-specific models, i.e.~models that can be deployed without incurring large computation or memory costs.
Several concurrent works have also proposed a similar idea to ours -- that of using extracted rationales as supervision~\citep{wang2022pinto,ho2022large,magister2022teaching,li2023symbolic}. Amongst them, PINTO~\citep{wang2022pinto} relies on an LLM to generate rationales at test-time, and thus does not fully solve deployment challenges. 
Compared with~\citet{ho2022large} and \citet{magister2022teaching}, 
we go beyond their experiments to provide a granular study by varying training dataset size, exploring downstream model sizes, and demonstrating the effectiveness of our method on fully unlabeled datasets.

\section{Distilling step-by-step}
We propose a new paradigm, \emph{\Method}, that leverages the ability of LLMs to reason about their predictions to train smaller models in a data-efficient way.
Our overall framework is illustrated in Figure~\ref{fig:framework}. Our paradigm has two simple steps: First, given an LLM and an unlabeled dataset, we prompt the LLM to generate output labels along with \textit{rationales} to justify the labels. Rationales are natural language explanations that provide support for the model's predicted label (see Figure~\ref{fig:framework}).
Second, we leverage these rationales in addition to the task labels to train smaller downstream models. 
Intuitively, rationales provide richer, more detailed information about why an input is mapped to a specific output label, and often contain relevant task knowledge that may be hard to infer solely from the original inputs.

\subsection{Extracting rationales from LLMs}
\label{sec:rationale-extraction}
Recent studies observe one intriguing emerging property of LLMs: their ability to generate rationales that support their predictions~\citep{wei2022chain,kojima2022large}. While the studies have largely focused on how to elicit such reasoning capability from LLMs~\citep{nye2021show,wei2022chain,kojima2022large}, we use them in training smaller downstream models.

\begin{figure}
    \centering
    \includegraphics[width=0.98\linewidth]{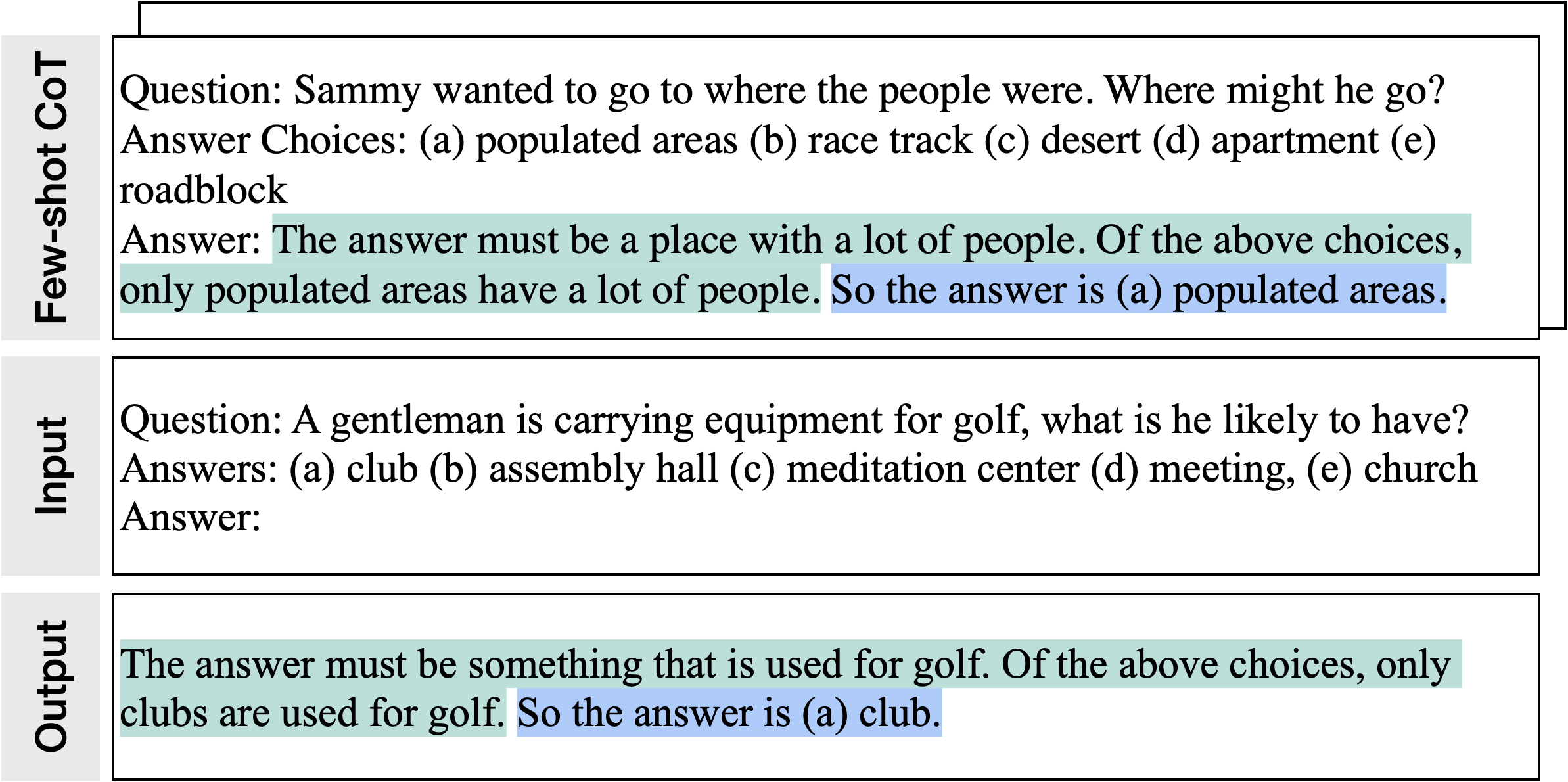}
    \vspace{-2mm}
    \caption{We use few-shot CoT prompting that contains both an example rationale (highlighted in green) and a label (highlighted in blue) to elicit rationales from an LLM on new input examples.}
    \label{fig:CoT-example}
    \vspace{-2mm}
\end{figure}

Specifically, we utilize Chain-of-Thought (CoT) prompting~\citep{wei2022chain} to elicit and extract rationales from LLMs.
As illustrated in Figure~\ref{fig:CoT-example}, given an unlabeled dataset $x_i \in D$, we first curate a prompt template $p$ that articulates how the task should be solved. Each prompt is a triplet $(x^{\mathrm{p}}, r^{\mathrm{p}}, y^{\mathrm{p}})$, where $x^{\mathrm{p}}$ is an example input, $y^{\mathrm{p}}$ is its corresponding label and $r^{\mathrm{p}}$ is a user-provided rationale that explains why $x^{\mathrm{p}}$ can be categorized as $y^{\mathrm{p}}$.
We append each input $x_i$ to $p$ and use it as an input to prompt the LLM to generate rationales and labels for each $x_i \in D$. With the demonstrations seen in $p$, the LLM is able to mimic the triplet demonstration to generate the rationale $\hat{r}_i$ and output $\hat{y}_i$ for $x_i$.

\subsection{Training smaller models with rationales}
\label{sec:multitask-training}
We first describe the current framework for learning task-specific models. With this framework in place, we extend it to  incorporate rationales into the training process.
Formally, we denote a dataset as $\mathcal{D} = \{(x_i, y_i)\}_{i=1}^N$ where each $x_i$ represents an input and $y_i$ is the corresponding desired output label. While our framework supports inputs and outputs of any modality, our experiments limits $x$ and $y$ to be natural language. This text-to-text framework~\citep{raffel2020exploring} encompasses a variety of NLP tasks: classification, natural language inference, question answering and more.

\paragraph{Standard finetuning and task distillation.}
The most common practice to train a task-specific model is to finetune a pretrained model with supervised data~\citep{howard-ruder-2018-universal}.
In the absence of human-annotated labels, task-specific distillation~\citep{hinton2015distilling,tang2019distilling} uses LLM teachers to generates pseudo noisy training labels, $\hat{y}_i$ in place of $y_i$~\citep{wang2021want,smith2022language,arora2022ask}.

For both scenarios, the smaller model $f$ is trained to minimize the label prediction loss:
\begin{align}
\label{eq:label_loss}
    \mathcal{L}_{\mathrm{label}} = \frac{1}{N} \sum_{i=1}^N \ell(f(x_i), \hat{y}_i),
\end{align}
where $\ell$ is the cross-entropy loss between the predicted and target tokens.
Note that for ease of exposition, we overload $\hat{y}_i$ in \eqref{eq:label_loss} to be either human-annotated labels $y_i$ for the standard finetuning case, or LLM-predicted labels $\hat{y}_i$ for the model distillation case.

\paragraph{Multi-task learning with rationales.}
To create a more explicit connection between $x_i$'s to $\hat{y}_i$'s, we use extracted rationales $\hat{r}_i$ as additional supervision.
There are several ways to incorporate rationales into the downstream model's training process. One straightforward approach is feed $\hat{r}_i$ as an additional input—as proposed by other concurrent research~\citep{rajani2019explain,wang2022pinto}.
In other words, the $f(x_i, \hat{r}_i) \rightarrow \hat{y}_i$ is trained with both text and rationale $[x, r]$ as inputs:
\begin{align}
\label{eq:pinto_loss}
    \mathcal{L} = \frac{1}{N} \sum_{i=1}^N \ell(f(x_i, \hat{r}_i), \hat{y}_i).
\end{align}
Unfortunately, this design requires an LLM to first generate a rationale before the $f$ can make a prediction. The LLM is still necessary during deployment, limited its deployability.

In this work, instead of using rationales as additional model inputs, we frame learning with rationales as a multi-task problem.
Specifically, we train the model $f(x_i) \rightarrow (\hat{y}_i, \hat{r}_i)$ to not only predict the task labels but also generate the corresponding rationales given the text inputs:
\begin{align}
\label{eq:distill-step-by-step}
    \mathcal{L} = \mathcal{L}_{\mathrm{label}} +  \lambda \mathcal{L}_{\mathrm{rationale}},
\end{align}
where $\mathcal{L}_{\mathrm{label}}$ is the label prediction loss in \eqref{eq:label_loss} and $\mathcal{L}_{\mathrm{rationale}}$ is the \textit{rationale generation loss}:
\begin{align}
    \mathcal{L}_{\mathrm{rationale}} = \frac{1}{N} \sum_{i=1}^N \ell(f(x_i), \hat{r}_i).
\end{align}
The rationale generation loss enables the model to learn to generate the intermediate reasoning steps for the prediction, and could therefore guide the model in better predicting the resultant label. This is our proposed \Method. Compared with~\eqref{eq:pinto_loss}, the rationale $\hat{r}_i$ is not required in the test time, which removes the need for an LLM at test-time.

We prepend ``task prefixes'' ($\texttt{[label]}$, $\texttt{[rationale]}$) to the input examples and train the smaller model to output $\hat{y}_i$ when $\texttt{[label]}$ is provided and to produce $\hat{r}_i$ with $\texttt{[rationale]}$~\citep{raffel2020exploring}.
\begin{figure*}[!t]
    \centering
    \includegraphics[width=0.99\linewidth]{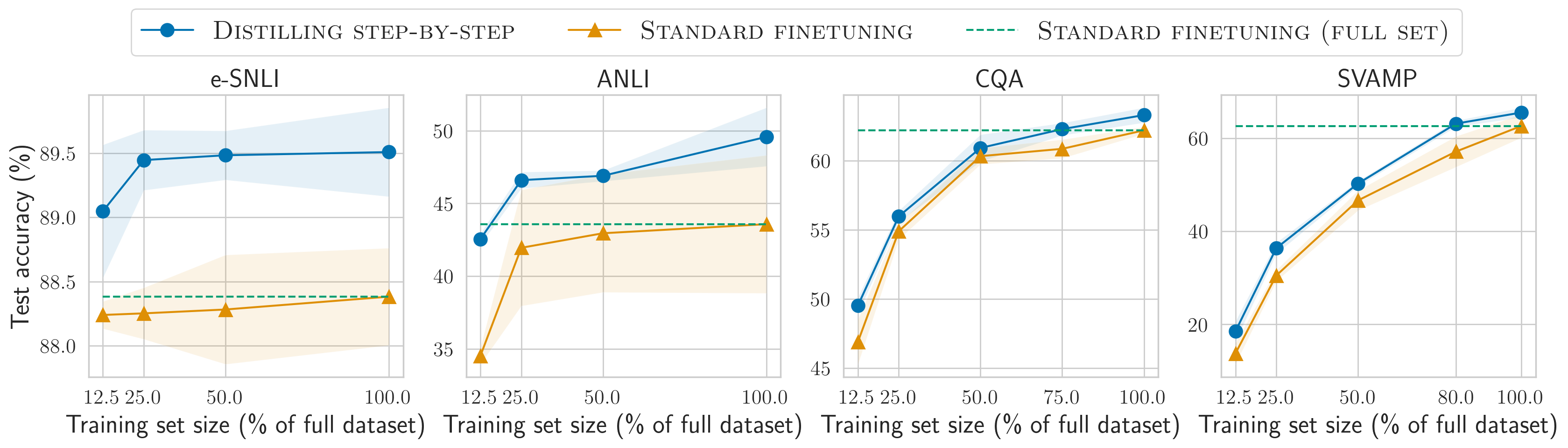}
    \vspace{-1mm}
    \caption{We compare \Method and Standard finetuning using $220$M T5 models on varying sizes of human-labeled datasets. On all datasets, \Method is able to outperform Standard finetuning, trained on the full dataset, by using much less training examples (e.g., $12.5\%$ of the full e-SNLI dataset).} 
    \label{fig:num_data_gt}
\end{figure*}
\begin{figure*}[!t]
    \centering
    \includegraphics[width=0.99\linewidth]{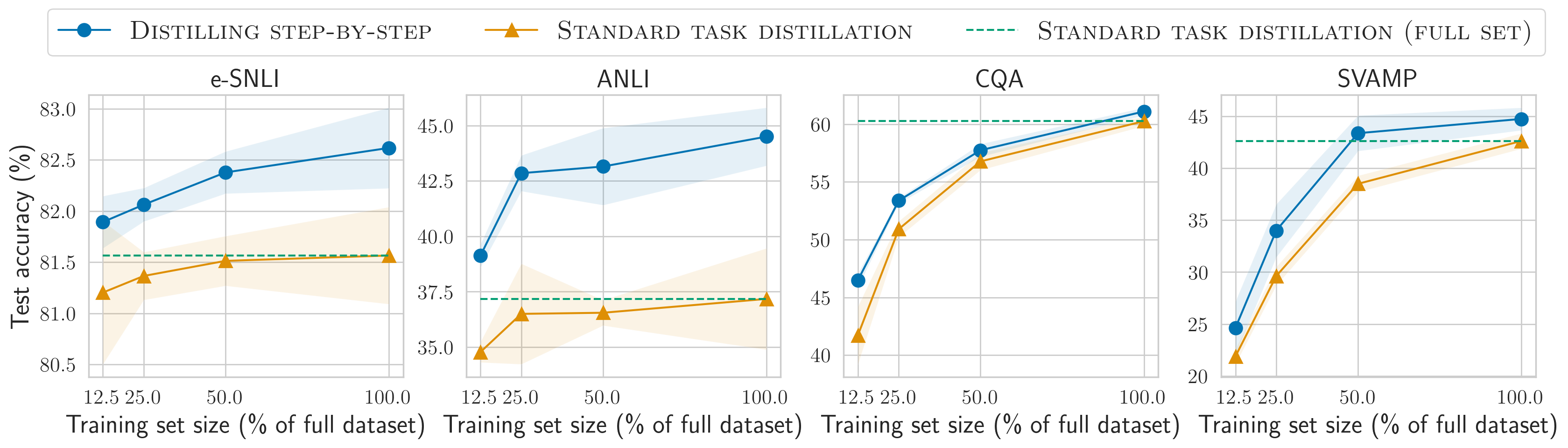}
    \vspace{-1mm}
    \caption{Similar to the plots above, we compare \Method and Standard task distillation using $220$M T5 models on varying sizes of unlabeled datasets. \Method is able to outperform Standard task distillation by using only a small subset of the full unlabeled dataset (e.g., $12.5\%$ on ANLI dataset).} 
    \label{fig:num_data_llm}
\end{figure*}

\section{Experiments}
We empirically validate the effectiveness of \Method. First, we show that when compared to standard finetuning and task distillation approaches, \Method achieves better performance with much fewer number of training examples, substantially improving the data efficiency to learn small task-specific models (Sec.~\ref{sec:exp-data-efficiency}).
Second, we show that \Method surpasses the performance of LLMs with much smaller model size, drastically lowering the deployment cost compared to LLMs (Sec.~\ref{sec:exp-deploy-efficiency}).
Third, we investigate the minimum resources required, w.r.t. both number of training examples and model size, for \Method to outperform LLMs. We show that \Method outperforms LLMs by using less data and smaller model, simultaneously improving both data- and deployability-efficiency (Sec.~\ref{sec:exp-data-model}).
Finally, we perform ablation studies to understand the influence of different components and design choices in the \Method framework (Sec.~\ref{sec:exp-ablation}).

\paragraph{Setup.}
In the experiments, we consider the $540$B PaLM model~\citep{chowdhery2022palm} as the LLM. For task-specific downstream models, we use T5 models~\citep{raffel2020exploring} where we initialize the models with pretrained weights obtained from publicly available sources\footnote{\url{https://huggingface.co/}}. 
For CoT prompting, we follow~\citet{wei2022chain} when available, and curate our own examples for new datasets. We include more implementation details in Appendix~\ref{sec:appendix-implementation}.

\paragraph{Datasets.}
We conduct the experiments on $4$ popular benchmark datasets across 3 different NLP tasks: \textit{e-SNLI}~\citep{camburu2018snli} and \textit{ANLI}~\citep{nie-etal-2020-adversarial} for natural language inference; \textit{CQA}~\citep{talmor-etal-2019-commonsenseqa,rajani2019explain} for commonsense question answering; \textit{SVAMP}~\citep{patel-etal-2021-nlp} for arithmetic math word problems. We include more dataset details in Appendix~\ref{sec:appendix-dataset}.

\subsection{Reducing training data}
\label{sec:exp-data-efficiency}

We compare \Method to two most common methods in learning task-specific models: (1) \textsc{Standard finetuning} when human-labeled examples are available, and (2) \textsc{Standard task distillation} when only unlabeled examples are available. Specifically, standard finetuning refers to the prevailing pretrain-then-finetune paradigm that finetunes a model with ground-truth labels via standard label supervision~\citep{howard-ruder-2018-universal}. On the other hand, when only unlabeled examples are available, standard task distillation learns the task-specific model by treating a teacher LLM's predicted labels as ground-truths~\citep{hinton2015distilling,chen2020big,wang2021want,smith2022language,arora2022ask}.

In the following set of experiments, we fix the task-specific models to be 220M T5-Base models, and compare the task performances achieved by different methods under varying number of available training examples.

\paragraph{\Method outperforms standard finetuning with much less labeled examples.}
When finetuned with human-labeled examples, Figure~\ref{fig:num_data_gt} shows that \Method consistently achieves better performance than standard finetuning across varying numbers of labeled examples used.
Furthermore, we see that \Method can achieve the same performance as standard finetuning with much less labeled examples. In particular, by using only $12.5\%$ of the full e-SNLI dataset, \Method can outperform standard finetuning trained with $100\%$ of the full dataset. Similarly, we achieve $75\%$, $25\%$, and $20\%$ reduction in training examples required to outperform standard finetuning on ANLI, CQA, and SVAMP respectively.

\paragraph{\Method outperforms standard distillation with much less unlabeled examples.}
When only unlabeled data is available, we compare \Method to standard task distillation. In Figure~\ref{fig:num_data_llm}, we observe an overall similar trend to the finetuning setup. Specifically, we see that \Method outperforms standard task distillation on all $4$ datasets under different numbers of unlabeled data used.
We as well see that \Method requires much less unlabeled data to outperform standard task distillation. For instance, we need only $12.5\%$ of the full unlabeled dataset to outperform the performance achieved by standard task distillation using $100\%$ of the training examples on e-SNLI dataset.

\subsection{Reducing model size}
\label{sec:exp-deploy-efficiency}

In the following set of experiments, we hold the training set size fixed (using $100\%$ of the datasets), and compare varying sizes of small T5 models trained with \Method and standard approaches to LLMs.
Specifically, we consider $3$ different sizes of T5 models, i.e., 220M T5-Base, 770M T5-Large, and 11B T5-XXL.
For LLMs, we include two baseline methods: (1) \textsc{Few-shot CoT}~\citep{wei2022chain}, and (2) \textsc{PINTO tuning}~\citep{wang2022pinto}. Few-shot CoT directly utilizes CoT demonstrations to prompt the 540B PaLM to generate intermediate steps before predicting the final labels without any further finetuning of the LLM. 
PINTO tuning refers to our extension of \citet{wang2022pinto} to handle tasks beyond question-answering, which are not studied by \citet{wang2022pinto}.
Here, we finetune a 220M T5-Base model on top of the outputs generated from the PaLM model, which can be viewed as a finetuning method for LLMs with additional parameters~\citep{zhang2020side,lester2021power}.

\begin{figure*}[!t]
    \centering
    \includegraphics[width=0.99\linewidth]{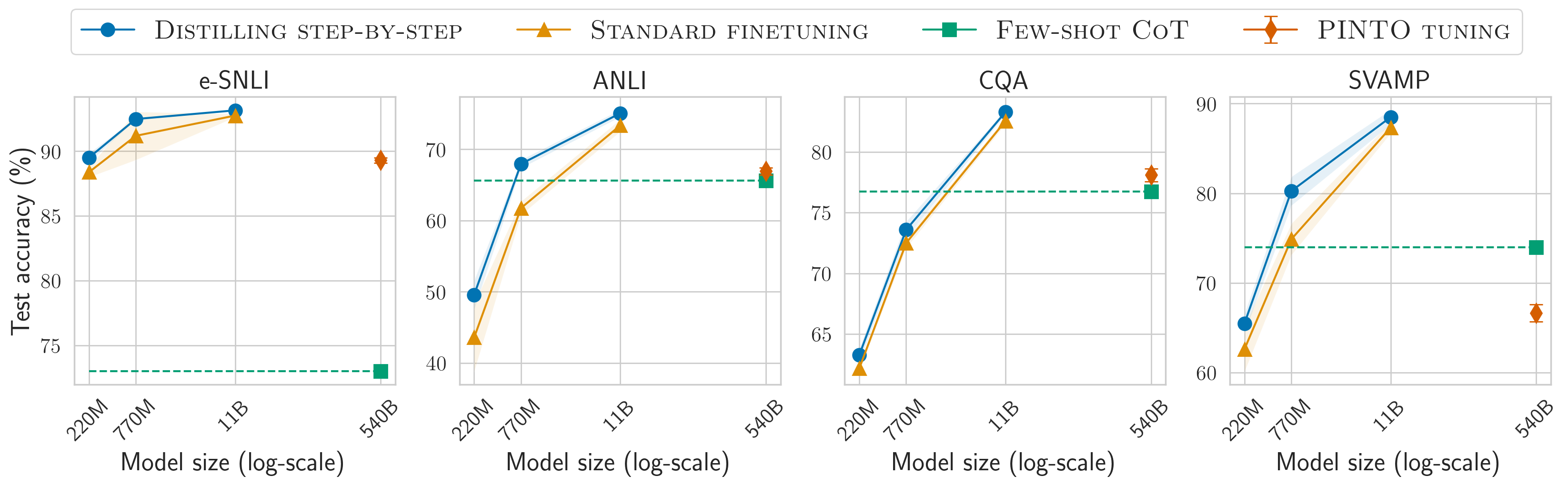}
    \vspace{-1mm}
    \caption{We perform \Method and Standard finetuning, using the full human-labeled datasets, on varying sizes of T5 models and compare their performance to LLM baselines, i.e., Few-shot CoT and PINTO Tuning. \Method is able to outperform LLM baselines by using much smaller models, e.g., over $700\times$ smaller model on ANLI. Standard finetuning fails to match LLM's performance using the same model size.}
    \label{fig:model_size_gt}
\end{figure*}

\begin{figure*}[!t]
    \centering
    \includegraphics[width=0.99\linewidth]{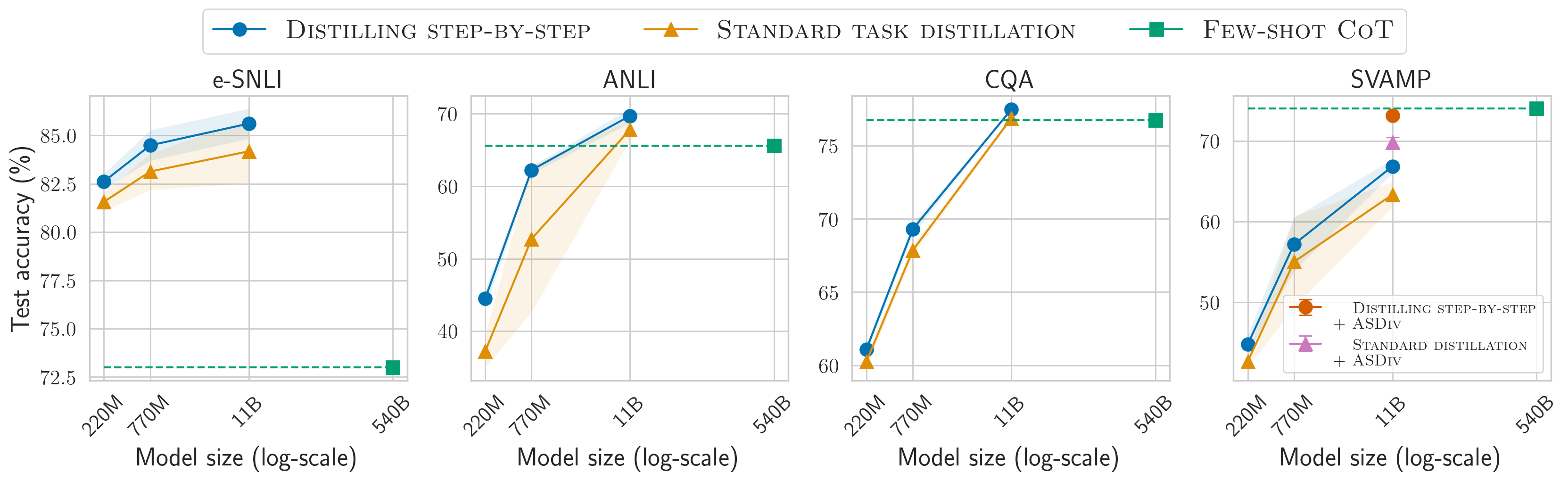}
    \vspace{-1mm}
    \caption{Using unlabeled datasets, we perform \Method and Standard task distillation on varying sizes of T5 models and compare them to Few-shot CoT. \Method outperforms Few-shot CoT by using $2000\times$ smaller models on e-SNLI and $45\times$ smaller models on ANLI and CQA. On SVAMP, by adding unlabeled examples from ASDiv, we close the gap to Few-shot CoT whereas Standard distillation still struggles to catch up.}
    \label{fig:model_size_llm}
\end{figure*}

We present the experimental results under the two broad scenarios of having access to labeled datasets or unlabeled datasets in Figure~\ref{fig:model_size_gt} and Figure~\ref{fig:model_size_llm}, respectively. We plot each method by their deployed model sizes for prediction ($x$-axis), and their corresponding task performances ($y$-axis).

\paragraph{\Method improves over standard baselines across varying model sizes used.}
In Figure~\ref{fig:model_size_gt} and Figure~\ref{fig:model_size_llm} respectively, we see that Distilling step-by-step consistently improves over standard finetuning and standard distillation across all sizes of T5 models.
The improvements are most pronounced on ANLI, where \Method outperforms standard finetuning and distillation by an average of $8\%$ and $13\%$ on task accuracy respectively.

\paragraph{\Method outperforms LLMs by using much smaller task-specific models.}
In Figure~\ref{fig:model_size_gt} when human-labeled datasets are available, \Method can always outperform Few-shot CoT and PINTO tuning on all $4$ datasets considered, by using much smaller T5 models. For instance, we can achieve better performances than 540B PaLM model's Few-shot CoT with $220$M (over $2000\times$ smaller) T5 model on e-SNLI, $770$M (over $700\times$ smaller) T5 models on ANLI and SVAMP, and $11$B (over $45\times$ smaller) T5 model on CQA. These results hold true even by further finetuning the 540B PaLM model on available labeled data with PINTO tuning\footnote{We note that PETuning methods may outperform PINTO tuning. However, they require massive resource in both training and deployment, which is not the focus of this work.}.

In Figure~\ref{fig:model_size_llm}, by only utilizing unlabeled examples, \Method also outperforms the teacher LLM on 3 out of 4 datasets. Specifically, \Method surpasses the $540$B PaLM model's Few-shot CoT performance by using $11$B T5 with less than $3\%$ of PaLM's size.
On SVAMP where the distilled model underperforms, we hypothesize that the performance gap is due to the relatively small number of data points in the dataset (i.e., $800$). In reaction, we propose to augment the dataset with additional unlabeled examples to close the performance gap as shown in next.

\paragraph{Unlabeled data augmentation further improves \Method.}
We augment the SVAMP training set with unlabeled examples from the \textit{ASDiv} dataset~\citep{miao-etal-2020-diverse}. ASDiv dataset contains a total of $2,305$ examples, where each example is a math word problem similar to the ones in SVAMP. In Figure~\ref{fig:model_size_llm} on SVAMP, we show the performances of \Method and standard task distillation using $11$B T5 model after augmenting the training set with ASDiv.
We see the data augmentation much improves the performance for both \Method and standard task distillation. However, even with the added unlabeled examples, standard task distillation still underperforms Few-shot CoT. On the other hand, \Method is able to much more efficiently exploit the value of the added examples to achieve the same performance level of Few-shot CoT, again, using a T5 model of size less than $3\%$ of the $540$B PaLM.

\begin{figure*}[!t]
    \centering
    \includegraphics[width=0.99\linewidth]{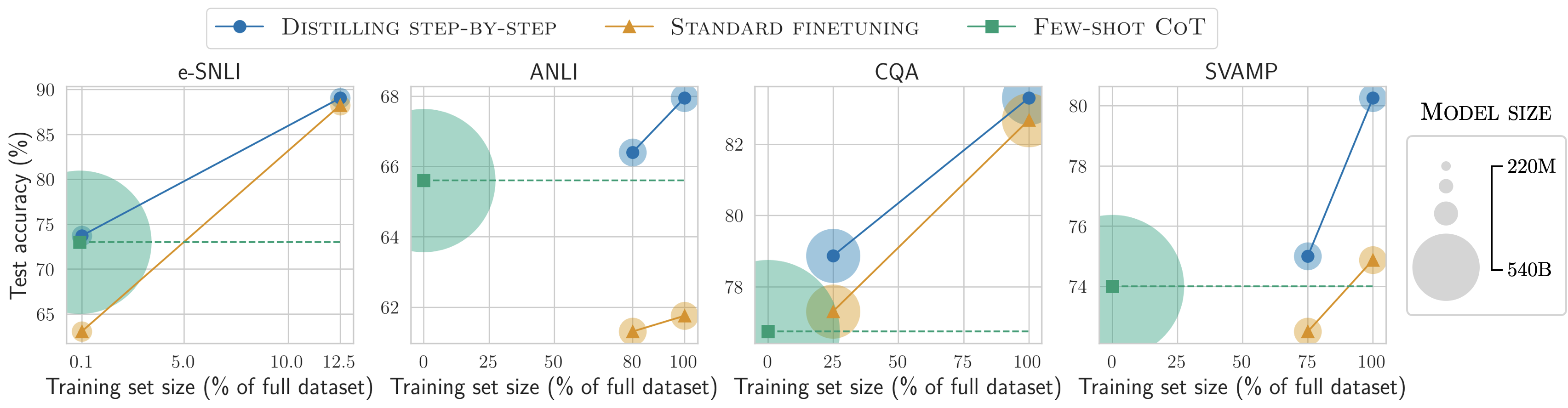}
    \vspace{-1mm}
    \caption{We show the minimum size of T5 models and the least amount of human-labeled examples required for \Method to outperform LLM's Few-shot CoT by a coarse-grained search. \Method is able to outperform Few-shot CoT using not only much smaller models, but it also achieves so with much less training examples compared to Standard finetuning. On ANLI, we outperform the LLM CoT with a $770$M model using only $80\%$ of the dataset, whereas Standard finetuning struggles to match even using $100\%$ of the dataset.}
    \label{fig:data_model_gt}
\end{figure*}

\begin{figure*}[!t]
    \centering
    \includegraphics[width=0.99\linewidth]{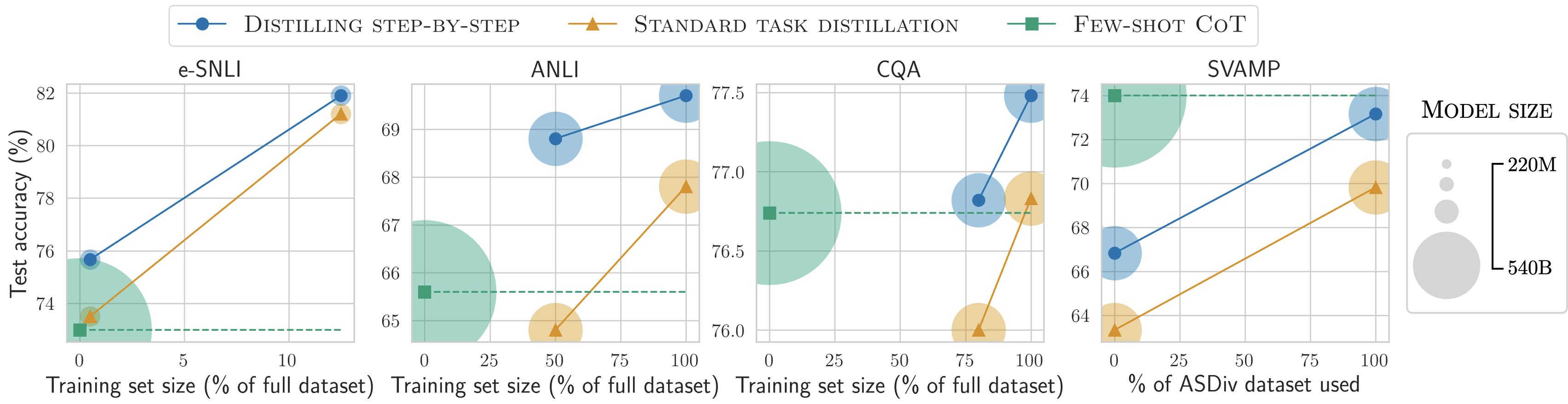}
    \vspace{-1mm}
    \caption{Similar to Figure~\ref{fig:data_model_gt} but using only unlabeled examples, \Method is able to outperform Few-shot CoT using much smaller models and with much less examples compared to Standard task distillation. On SVAMP, the $x$-axis corresponds to the size of ASDiv dataset used for augmenting the original SVAMP dataset, i.e., $x=0$ is without augmentation and $x=100$ corresponds to adding the full ASDiv dataset.}
    \label{fig:data_model_llm}
\end{figure*}

\subsection{Outperforming LLMs using minimum model size and least training data}
\label{sec:exp-data-model}

Here, using the LLM's performance as an anchor point, we explore the most efficient resource requirements in terms of both number of training examples and deployed model size, that \Method and standard finetuning/distillation need to outperform the LLM.
We present the results, again under human-labeled setting and unlabeled setting, in Figure~\ref{fig:data_model_gt} and Figure~\ref{fig:data_model_llm} respectively.
We visualize the results by plotting different resultant models by (1) the number of training examples used ($x$-axis), (2) the final task performance achieved ($y$-axis), and (3) the size of the model (visualized by the size of the shaded area).

\paragraph{\Method outperforms LLMs with much smaller models by using less data.}
On all datasets in Figure~\ref{fig:data_model_gt}, we see that \Method outperforms PaLM's Few-shot CoT with much smaller T5 models using only a subset of the available training examples. Specifically, on e-SNLI, \Method can achieve better performance than Few-shot CoT with a model over $2000\times$ smaller (220M T5) and only $0.1\%$ of the full dataset. In Figure~\ref{fig:data_model_llm} where only unlabeled datasets are available, we observe the same trend that \Method can, at most time, outperform Few-shot CoT with smaller model as well as less data. For instance, on ANLI, \Method outperforms the LLM with a $45\times$ smaller model and $50\%$ of the full unlabeled set.

\paragraph{Standard finetuning and distillation require more data and larger model.}
Finally, in Figure~\ref{fig:data_model_gt} and Figure~\ref{fig:data_model_llm}, we see that standard finetuning and distillation often need either more data or larger models to match LLM's performance. For instance, on e-SNLI in Figure~\ref{fig:data_model_gt}, we observe that \Method outperform the LLM using only $0.1\%$ of the dataset while standard finetuning requires more data to match the performance.
Furthermore, on ANLI in Figure~\ref{fig:data_model_gt}, we observe that \Method can outperform PaLM using $770$M model with only $80\%$ of the training set while standard finetuning struggles to match the LLM even using the full dataset and thus requires larger model to close the performance gap.

\subsection{Further ablation studies}
\label{sec:exp-ablation}

So far, we have focused on showing the effectiveness of \Method on reducing the training data required for finetuning or distilling smaller task-specific models. In this section, we perform further studies to understand the influence of different components in the \Method framework. Specifically, we study (1) how different LLMs, from which the rationales are extracted, affect the effectiveness of \Method, and (2) how the multi-task training approach compares to other potential design choices in training small task-specific models with LLM rationales.
Here, we fix the small task-specific models to be $220$M T5 models, and utilize $100\%$ of the data on all datasets.

\paragraph{\Method works with different sizes of decently trained LLMs.}
In addition to using $540$B PaLM as the LLM, here we consider a relatively smaller LLM, $20$B GPT-NeoX model~\cite{gpt-neox-20b}, from which we extract rationales for \Method.
In Table~\ref{table:small-teacher}, we see that when coupled with LLMs of different sizes, \Method can still provide performance improvements compared to standard finetuning. However, the performance lift is smaller when rationales are extracted from the $20$B GPT-NeoX model instead of from the $540$B PaLM. This can be due to the fact that the larger PaLM model provides higher-quality rationales that are more beneficial for learning the task.

\begin{table}[!t]
\caption{\Method works with different sizes of LLMs. When rationales are extracted from a $20$B GPT-NeoX model, \Method is still able to provide performance lift compared to standard finetuning on $220$M T5 models.}
\label{table:small-teacher}
\small
\centering
\begin{adjustbox}{width=1\linewidth}
\begin{tabular}{llcccc}
\toprule
& & \multicolumn{4}{c}{Dataset} \\
\cmidrule{3-6}
Method & LLM &  e-SNLI & ANLI & CQA & SVAMP\\
\midrule
\textsc{Standard finetuning} & N/A & 88.38 & 43.58 & 62.19 & 62.63 \\
\textsc{\Method} & $20$B & 89.12 & 48.15 & 63.25 & 63.00 \\
\textsc{\Method} & $540$B & 89.51 & 49.58 & 63.29 & 65.50 \\
\bottomrule
\end{tabular}
\end{adjustbox}
\end{table}

\paragraph{Multi-task training is much more effective than single-task rationale and label joint prediction.}
There are different possible ways to train task-specific models with LLM-rationales as output supervisions. One straightforward approach is to concatenate the rationale $\hat{r}_i$ and label $\hat{y}_i$ into a single sequence $[\hat{r}_i, \hat{y}_i]$ and treat the entire sequence as the target output in training small models, as considered in~\cite{magister2022teaching,ho2022large}:
\begin{align}
    \mathcal{L}_{\mathrm{single}} = \frac{1}{N} \sum_{i=1}^N \ell(f(x_i), [\hat{r}_i, \hat{y}_i]).
\end{align}
In Table~\ref{table:multi-task}, we compare this single-task training approach to our proposed multi-task training approach for utilizing LLM-rationales. We see that not only multi-task training consistently leads to better performance, single-task training with LLM-rationales can at times leads to worse performance than standard finetuning, e.g., on ANLI and CQA.
In fact, similar results have also been observed in~\cite{wiegreffe-etal-2021-measuring,magister2022teaching,ho2022large} that simply treating rationale and label predictions as a single joint task may harm the model's performance on label prediction.
This validates our use of the multi-task training approach, and highlights the need to treat the rationales carefully so as to unleash their actual benefits.

\begin{table}[!t]
\caption{Our proposed multi-task training framework consistently leads to better performances than treating rationale and label predictions as a single task. Single-task training can at times lead to worse performance than standard finetuning.}
\label{table:multi-task}
\small
\centering
\begin{adjustbox}{width=1\linewidth}
\begin{tabular}{lcccc}
\toprule
& \multicolumn{4}{c}{Dataset} \\
\cmidrule{2-5}
Method &  e-SNLI & ANLI & CQA & SVAMP\\
\midrule
\textsc{Standard finetuning} & 88.38 & 43.58 & 62.19 & 62.63 \\
\textsc{Single-task training} & 88.88 & 43.50 & 61.37 & 63.00 \\
\textsc{Multi-task training} & \textbf{89.51} & \textbf{49.58} & \textbf{63.29} & \textbf{65.50} \\
\midrule
\end{tabular}
\end{adjustbox}
\end{table}
\section{Discussion}
We propose \Method to extract rationales from LLMs as informative supervision in training small task-specific models. We show that \Method reduces the training dataset required to curate task-specific smaller models; it also reduces the model size required to achieve, and even surpass, the original LLM's performance. \Method proposes a resource-efficient training-to-deployment paradigm compared to existing methods. Further studies demonstrate the generalizability and the design choices made in \Method. Finally, we discuss the limitations, future directions and ethics statement of our work below.

\section*{Limitations}
There are a number of limitations with our approach. First, we require users to produce a few example demonstrations ($\sim 10$-shot for all tasks) in order to use the few-shot CoT~\citep{wei2022chain} prompting mechanism. This limitation can be improved by using recent advances that suggest that rationales can be elicited without any user-annotated demonstrations~\citep{kojima2022large}.
Second, training task-specific models with rationales incur slight training-time computation overhead. However, at test time, our multi-task design naturally avoids the computation overhead since it allows one to only predict labels without generating the rationales.
Finally, while we observe success using LLM rationales, there is evidence that LLMs exhibit limited reasoning capability on more complex reasoning and planning tasks~\citep{valmeekam2022large}. Future work should characterize how rationale quality affects \Method.

\section*{Ethics statement}
It is worth noting that the behavior of the our downstream smaller models is subject to biases inherited from the larger teacher LLM.
We envision that the same research progress in reducing anti-social behaviors in LLMs can also be applied to improve  smaller language models.

\bibliographystyle{acl_natbib}
\bibliography{references}

\clearpage
\newpage

\appendix
\section{Experiment detail}
\subsection{Implementation}
\label{sec:appendix-implementation}
We perform our experiments on cloud A100$\times$16 GPU instances. We train the T5 models with the following hyperparameters, using publicly available packages from \url{https://github.com/huggingface/transformers}:
\begin{itemize}
    \item T5-Base ($220$M) and T5-Large ($770$M): We train the models with $\textrm{learning rate} = 5 \times 10^{-5}$, $\textrm{batch size} = 64$, $\textrm{max input length} = 1024$, for a maximum of $10000$ steps.
    
    \item T5-XXL ($11$B): We train the models with $\textrm{learning rate} = 5 \times 10^{-5}$, $\textrm{batch size} = 32$, $\textrm{max input length} = 1024$, for a maximum of $4000$ steps.
\end{itemize}
We report all the results over $4$ random runs, and include the standard error in the presented plots.

\subsection{Datasets}
\label{sec:appendix-dataset}
We provide more detailed descriptions on the datasets used in our experiments. We include the sources from which we obtain the datasets as well as their original sources released from the authors. We refer readers to these sources for their license or terms for use and/or distribution. To the best of our knowledge, the datasets used do not contain information that names or uniquely identifies individual people or offensive content.
\begin{itemize}
    \item e-SNLI: The dataset was originally released in~\citep{camburu2018snli}, and made publicly available at \url{https://github.com/OanaMariaCamburu/e-SNLI}. We obtain the dataset from \url{https://huggingface.co/datasets/esnli}.
    \item ANLI: The dataset was originally released in~\citep{nie-etal-2020-adversarial}, and made publicly available at \url{https://github.com/facebookresearch/anli}. We obtain the dataset from \url{https://huggingface.co/datasets/anli}. We use the R1 split in our experiments.
    \item CQA: The dataset was originally released in~\citep{talmor-etal-2019-commonsenseqa}, and made publicly available at \url{https://www.tau-nlp.sites.tau.ac.il/commonsenseqa}. It was then augmented with human-labeled explanations by~\citep{rajani2019explain}, which is available at \url{https://github.com/salesforce/cos-e}. We obtain the dataset used in our experiments from \url{https://huggingface.co/datasets/cos_e}.
    \item SVAMP: The dataset was originally released in~\citep{patel-etal-2021-nlp}. We obtain the dataset from \url{https://github.com/arkilpatel/SVAMP}.
    \item ASDiv: The dataset was originally released in~\citep{miao-etal-2020-diverse}. We obtain the dataset from \url{https://github.com/chaochun/nlu-asdiv-dataset}.
\end{itemize}

For each dataset, we randomly subsample $10\%$ of the original training set to serve as validation set when validation set is not originally provided. For CQA, we use the original validation set to serve as our test set since the ground-truth labels are not available for the original test set. We provide the dataset statistics in Table~\ref{table:dataset}.

\begin{table}[!t]
\caption{Dataset statistics used in our experiments.}
\label{table:dataset}
\small
\centering
\begin{tabular}{lccccc}
\toprule
 
Dataset & Train & Validation & Test \\
\midrule
e-SNLI & 549,367 & 9,842 & 9,824 \\
ANLI &  16,946 & 1,000 & 1,000 \\
CQA  & 8,766 & 975 & 1,221\\
SVAMP & 720 & 80 & 200 \\
\bottomrule
\end{tabular}
\end{table}

\end{document}